%% file: access.tex
\documentclass{ieeeaccess}

\usepackage{cite}
\usepackage{amsmath,amssymb,amsfonts}
\usepackage{algorithmic}
\usepackage{graphicx}
\usepackage{textcomp}
\usepackage{booktabs}

\usepackage{threeparttable}
\usepackage{multirow}
\usepackage[hyphens]{url}

\usepackage{hyperref}

\usepackage{cleveref}

\usepackage {tablefootnote}

\usepackage{amsmath,amsfonts,amssymb,bm}

\usepackage{caption, subcaption}
\def\BibTeX{{\rm B\kern-.05em{\sc i\kern-.025em b}\kern-.08em
    T\kern-.1667em\lower.7ex\hbox{E}\kern-.125emX}}
    
\begin{document}
\history{Received 10 October 2024, accepted 20 December 2024, date of publication 25 December 2024, date of current version 15 January 2025.}
\doi{10.1109/ACCESS.2024.3522972}

\title{
    Majority or Minority: Data Imbalance Learning Method for Named Entity Recognition
}

\author{
    \uppercase{SOTA NEMOTO}\authorrefmark{1}, \uppercase{SHUNSUKE KITADA}\authorrefmark{1}, and \uppercase{HITOSHI IYATOMI}\authorrefmark{1},
    \IEEEmembership{Member, IEEE}
}

\address[1]{
    Department of Applied Informatics, Graduate School of Science and Engineering, Hosei University Tokyo, Japan \\
    (e-mail: \{sota.nemoto.5s, shunsuke.kitada.0831\}@gmail.com, iyatomi@hosei.ac.jp)
}

\markboth
{Nemoto \headeretal: Majority or Minority: Data Imbalance Learning Method for Named Entity Recognition}
{Nemoto \headeretal: Majority or Minority: Data Imbalance Learning Method for Named Entity Recognition}

\corresp{Corresponding author: SOTA NEMOTO (e-mail: sota.nemoto.5s@gmail.com).}

\begin{abstract}
\input{00_abstract}
\end{abstract}


\titlepgskip=-21pt

\maketitle

\section{Introduction}\label{sec:introduction}
\input{01_introduction}

\section{Related Work}\label{sec:related_work}
\input{02_related_work}

\section{Proposed Method}\label{sec:proposed_method}
\input{03_proposed_strategy}

\section{Experiments}\label{sec:experiments}
\input{04_experiments}

\section{Results}\label{sec:results}
\input{05_results}

\section{Discussion}\label{sec:discussion}
\input{06_discussion}

\section{Conclusion}\label{sec:conclusion}
\input{07_conclusion}

\bibliographystyle{IEEEtran}
\bibliography{reference}

\begin{IEEEbiography}[{\includegraphics[width=1in,height=1.25in,clip,keepaspectratio]{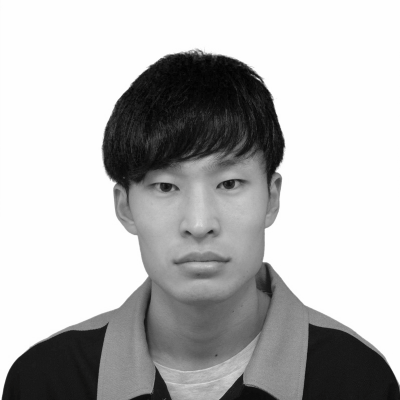}}]{Sota Nemoto} 
received the B.E. degree in engineering from Hosei University, Tokyo, Japan, in 2019, where he is currently pursuing the M.E. degree.
His research interests include information extraction and analysis of free-writing text of natural language processing.
He was a recipient of the Sponsor Honorable Mention from the Natural Language Processing Studies of Japan, in 2024.
\end{IEEEbiography}

\begin{IEEEbiography}[{\includegraphics[width=1in,height=1.25in,clip,keepaspectratio]{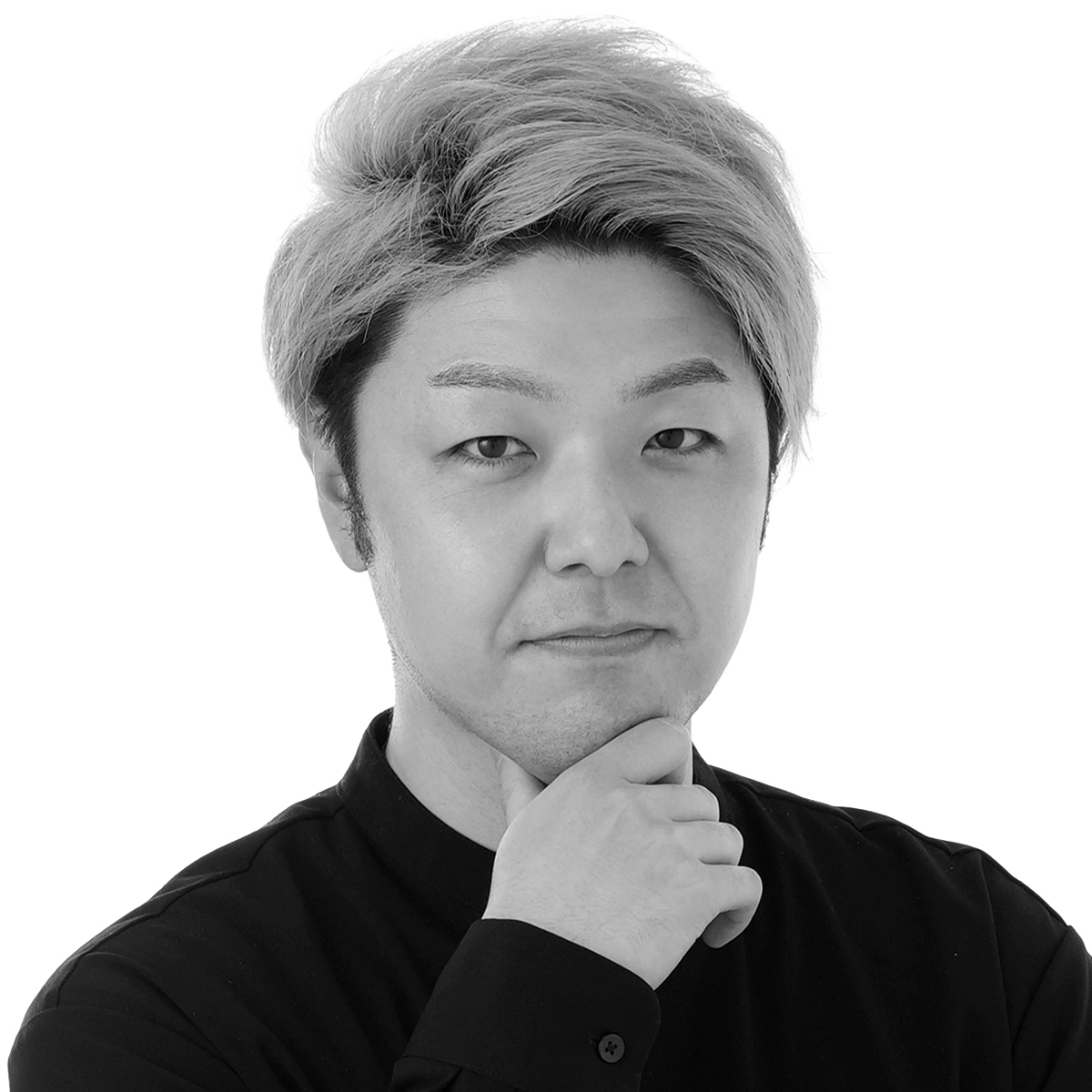}}]{Shunsuke Kitada}
received the B.E., M.E., and Ph.D. degrees in engineering from Hosei University, Tokyo, Japan, in 2018, 2020, and 2023, respectively.
He was a Japan Society for the Promotion of Science (JSPS) Research Fellow DC2 with Hosei University.
He is currently a Research Scientist with LYCorp and a Project Researcher with Hosei University.
His research interests include natural language processing, computer vision, and computational advertising.
He was a recipient of the Student Honorable Mention from the Information Processing Society of Japan, in 2019, and the Honorable Mention from the Young Researcher Association for NLP Studies, in 2019.
\end{IEEEbiography}

\begin{IEEEbiography}[{\includegraphics[width=1in,height=1.25in,clip,keepaspectratio]{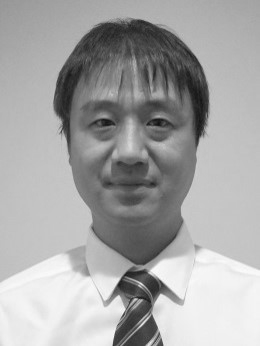}}]{Hitoshi Iyatomi}
received the B.E., M.E., and Ph.D. degrees in engineering from Keio University, Japan, in 1998, 2000, and 2004, respectively, and the Ph.D. degree in medical science from Tokyo Women’s Medical University, in 2011.
From 2000 to 2004, he worked for Hewlett-Packard Japan as a Technical Consultant.
He joined Hosei University, Japan, as a Research Associate, in 2004. From 2016 to 2017, he was a Visiting Scholar with Johns Hopkins University.
He is currently a Professor of applied informatics with Hosei University.
He has authored and co-authored more than 160 peer-reviewed journal and conference papers in various research areas based on machine learning, such as computer vision, medical applications, and natural language processing.
\end{IEEEbiography}

\EOD

\end{document}

%% file: 00_abstract.tex
Data imbalance presents a significant challenge in various machine learning (ML) tasks, particularly named entity recognition (NER) within natural language processing (NLP).
NER exhibits a data imbalance with a long-tail distribution, featuring numerous minority classes (i.e., entity classes) and a single majority class (i.e., $\mathcal{O}$-class).
This imbalance leads to misclassifications of the entity classes as the $\mathcal{O}$-class.
To tackle this issue, we propose a simple and effective learning method named majority or minority (MoM) learning.
MoM learning incorporates the loss computed only for samples whose ground truth is the majority class into the loss of the conventional ML model.
Evaluation experiments on four NER datasets (Japanese and English) showed that MoM learning improves the prediction performance of the minority classes without sacrificing the performance of the majority class and is more effective than widely known and state-of-the-art methods.
We also evaluated MoM learning using frameworks such as sequential labeling and machine reading comprehension, which are commonly used in NER.
Furthermore, MoM learning has achieved consistent performance improvements regardless of language, model, framework, or data size.

%% file: 01_introduction.tex
Named entity recognition (NER)~\cite{nadeau2007survey, lample2016neural} is one of many real-world natural language processing (NLP) tasks with significant data imbalance, especially when applied for business purposes like corporate information-gathering websites~\cite{guo2009named} and extracting drug names and diseases from vast amounts of unstructured medical data~\cite{ramachandran2021named}.
NER commonly uses a sequential labeling framework, a form of multiclass classification that predicts labels corresponding to the words in a sentence.
In sequential labeling, all words are divided into either entity words with information (i.e., proper nouns) or non-entity words without information.
Each entity word is labeled as a specific class (\texttt{PERSON}, \texttt{LOCATION}, etc.) to which a few samples belong.
In contrast, all non-entity words constitute the majority and are labeled as a single class (i.e., the ``others'' $\mathcal{O}$-class).
This labeling yields a data imbalance with a long-tail distribution.
Between the well-known benchmarks CoNLL2003~\cite{sang2003introduction} and OntoNotes5.0~\cite{pradhan2013towards}, the number of samples for the $\mathcal{O}$-class significantly exceeds that of the entity class, a condition that often leads to misclassifications of entity classes as the $\mathcal{O}$-class, causing a considerable decline in the prediction performance of the minority classes.
Overall, overcoming this data imbalance is a crucial step toward enhancing NER performance.

Conventional machine learning (ML) methods for addressing data imbalances are categorized into sampling-based methods~\cite{pouyanfar2018dynamic, buda2018systematic} for inputs and cost-sensitive learning~\cite{adel2017ranking, madabushi2019cost, li2020syrapropa} for outputs.
The sampling-based method, which adjusts the number of sentences in training, has a certain effect on the ML tasks.
However, NER uses sequential labeling, which predicts the labels corresponding to each word in a sentence; thus, it does not mitigate the imbalance.
By contrast, cost-sensitive learning addresses the imbalance by designing a loss function for the ML model based on the number of samples in each class.
While it is effective for binary classification, NER is a multiclass classification requiring an extension of this method.
This extension will lead to complex weight adjustments for each class and cases in which it is not fully capable, thus not attaining the desired level of performance.

In this paper, we propose a novel learning method, majority or minority (MoM) learning, to tackle the data imbalance in NER.
MoM learning is simple and effective for incorporating the loss computed only for samples whose ground truth is the majority class into the loss of the conventional ML model.
Our strategy enables cost-sensitive learning but differs from the concepts of previous studies because it does not depend on the difficulty of the classification or the number of samples in the class.
The purpose of MoM learning is to enhance performance by preventing misclassifications of the minority classes (entity classes) as the majority class (the $\mathcal{O}$-class). 
When incorporating the loss of entity classes instead of the $\mathcal{O}$-class, the model cannot distinguish whether the prediction is misclassified as the $\mathcal{O}$-class or as another entity class.
Therefore, MoM learning focuses on the $\mathcal{O}$-class to recognize misclassifications from the $\mathcal{O}$-class to the entity classes.

We evaluated MoM learning using four NER datasets~\cite{sang2003introduction, pradhan2013towards, hangyo2012building, ohmi2021ner} and with an ML model, including BERT~\cite{devlin2019bert}, which have proven successful in various NLP tasks.
The evaluation results demonstrated that MoM learning consistently improves performance across languages.
We also confirmed that MoM learning is more effective than those introduced in previous state-of-the-art studies, such as focal loss (FL)~\cite{lin2017focal} and dice loss (DL)~\cite{li2020dice}.
Furthermore, beyond common sequential labeling, we demonstrated the effectiveness of MoM learning using the machine reading comprehension (MRC) framework, which is becoming mainstream~\cite{li2020unified, zhang2023finbert}.

We summarize the contributions of this study as follows:
\begin{itemize}
    \item We propose a novel learning method, named majority or minority (MoM) learning, designed to address data imbalances with a long-tail distribution, a significant challenge in ML tasks.
    \item We evaluated four commonly used NER datasets (Japanese and English) and demonstrated that MoM learning is more effective than conventional methods in the sequential labeling framework and MRC.
    \item MoM learning improved the performance of the entity classes without compromising the performance of the $\mathcal{O}$-class in language-, model-, framework- and data size-agnostic.
\end{itemize}

The remainder of this paper is organized as follows.
\Cref{sec:related_work} reviews related work about the sampling-based method and const-sensitive learning.
\Cref{sec:proposed_method} describes the proposed method, MoM learning.
\Cref{sec:experiments} presents the evaluation experiments, including sequential labeling and MRC framework, followed by the datasets, loss functions, and implementation details.
\Cref{sec:results} shows the experimental results, showing that the proposed method consistently improves prediction performance for a wide range of NER tasks.
\Cref{sec:discussion} provides discussions of imbalances in NER, differences from conventional weighted losses, and results for non-sequences.
Finally, \cref{sec:conclusion} concludes this study.

%% file: 02_related_work.tex

\subsection{Sampling-based Method}

Standard sampling methods, such as random oversampling (ROS) and random undersampling (RUS)~\cite{pouyanfar2018dynamic, mikolov2013distributed}, are often used as a first choice.
ROS duplicates randomly selected minority samples, whereas RUS utilizes only selected samples for training. 
These methods enhance the impact of the minority samples and fix the imbalance before training.
Potential concerns with these methods are that ROS may lead to overfitting and increase the training time, whereas RUS may discard potentially effective training samples.


To address the limitations of these methods, various derived techniques have been proposed. For instance, the majority weighted minority oversampling technique (MWMOTE)\cite{barua2012mwmote} assigns weights to minority samples based on learning difficulty and processes them accordingly. Meanwhile, cluster-based majority under-sampling prediction (CBMP)\cite{zhang2010cluster} retrains only the samples contributing to learning in the majority class. However, since NER involves sequential labeling, adjusting the number of sentences in training through these methods does not improve the word-related imbalance. As these methods are unsuitable, especially in NER, where the imbalance is more severe than in other NLP tasks, we developed a new approach based on cost-sensitive learning.

\subsection{Cost-Sensitive Learning}

Weighted cross-entropy (WCE) is commonly employed to address the data imbalance by assigning class-specific weights to the cross-entropy loss based on the number of samples in each class~\cite{adel2017ranking, madabushi2019cost, li2020syrapropa}.
This method typically increases the weights of classes with few samples and decreases the weights of classes with many samples.
Because the distribution based on samples using training and tests only sometimes matches, WCE is unlikely to achieve the desired performance.

To overcome the limitations in WCE, FL~\cite{lin2017focal} and DL~\cite{li2020dice} have demonstrated superior effectiveness in binary classification compared to WCE~\cite{tran2021improving}, and these methods have been proven effective in NLP tasks.
These methods have been confirmed applicable to multiclass classification models, including extensions such as one-versus-the-rest and one-versus-one, as seen in conventional ML models (e.g., support vector machines and logistic regression).
For instance, FL is a loss function that addresses the imbalance between an image's background and target in a computer vision object detection task (i.e., binary classification).
FL adjusts weights by applying a power of $(1-p)$ to the predicted probability $p$ for each class.
Conversely, DL defines loss based on the dice coefficient~\cite{shamir2019continuous} (synonymous with the F1 score), which is the harmonic mean of precision and recall.
To maximize this dice coefficient, the number 1 minus it is defined as the loss function.
These methods differ from WCE in that they allow weights to be set based on the difficulty of classification rather than the number of samples per class.
As a result, the ML model minimizes the loss, enabling accurate classifications and increasing the loss for misclassifications.

Our method is based on cost-sensitive learning, and it addresses data imbalances by considering losses separately for the groups in which the most imbalances occur (i.e., majority and minority groups).
When focusing on the groups, MoM learning significantly differs from the concept of adding weights based on the difficulty in classifying, such as FL and DL.
In addition, MoM learning does not require weights to be set for each class, as in WCE.

%% file: 03_proposed_strategy.tex
This section describes the proposed method, MoM learning, which was MoM learning is designed for NER to improve the performance of ML models for data imbalance with a long-tail distribution characterized by the many minority classes and one majority class (i.e., entity classes and $\mathcal{O}$-class).

\subsection{Notation}

First, as a common approach to NER, we introduce the notation for sequential labeling.
We consider a dataset comprising a set of input sentences $\bm{X} = [\bm{x}^{(1)}, \cdots, \bm{x}^{(n)}, \cdots, \bm{x}^{(N)}]$ and the corresponding labels $\bm{Y} = [\bm{y}^{(1)}, \cdots, \bm{y}^{(n)}, \cdots, \bm{y}^{(N)}]$, where $N$ is the number of sentences in the dataset.
The $n$-th sentence split tokens and corresponding labels are represented as $\bm{x}^{(n)} = [\bm{w}^{(n)}_1, \cdots, \bm{w}^{(n)}_i, \cdots, \bm{w}^{(n)}_M]$ and $\bm{y}^{(n)} = [\bm{y}^{(n)}_1, \cdots, \bm{y}^{(n)}_i, \cdots, \bm{y}^{(n)}_M]$, respectively.
$M$ is the number of tokens in the longest sentence in the dataset.

The training label $\bm{y}^{(n)}_i$ is annotated using the BIO format~\cite{ramshaw1999text} in sequential labeling.
This format consists of entity classes (e.g., \texttt{PER}, \texttt{LOC}, and \texttt{ORG}) and a non-entity class (i.e., the $\mathcal{O}$-class), where the former is represented by prefixing the entity category with \texttt{B} for the first token and \texttt{I} for the rest, as follows: \texttt{B-PER}, \texttt{I-PER}, \texttt{B-LOC}, etc.
The sequence of predicted labels is denoted as $\bm{p}^{(n)} = [\bm{p}^{(n)}_1, \cdots, \bm{p}^{(n)}_i, \cdots, \bm{p}^{(n)}_M]$, where the ML model estimates the predicted probabilities $\bm{p}^{(n)}$ for each token of the sentence $\bm{x}^{(n)}$.

\subsection{Majority or Minority (MoM) Learning}

\input{figures/tex/ponti}

MoM learning is a simple and effective method that incorporates the loss for samples whose ground truth is a single majority class into the loss of an arbitrary conventional ML model.
\Cref{fig:ponti} illustrates the concept of MoM learning, where conventional loss $\mathcal{L}$ represents any loss function of the model, such as cross-entropy, which computes the loss for all samples boxed in red.
The $\mathcal{L}_{\mathrm{MoM}}$ term only computes the loss whose ground truth is the $\mathcal{O}$-class (i.e., $\bm{y}^{(n)}_i=\text{``}\mathcal{O}\text{''}$) framed in red, and incorporates them into the conventional loss.
The equation for the $\mathcal{L}_{\mathrm{MoM}}$ is presented as follows:
\begin{equation}
    \mathcal{L}_{\mathrm{MoM}}(\bm{y}^{(n)}, \bm{p}^{(n)}) = - \frac{1}{M}  \sum_{\bm{y}^{(n)}_{i} = \text{``}\mathcal{O}\text{''}}^{M} \ell(\bm{y}^{(n)}_{i}, \bm{p}^{(n)}_{i}),
\end{equation}
where $\ell$ is an arbitrary loss function, including cross-entropy, weighted cross-entropy, FL~\cite{lin2017focal}, DL~\cite{li2020dice}, etc.

Because $\mathcal{L}_{\mathrm{MoM}}$ focuses only on the $\mathcal{O}$-class, certain entity classes misclassified by the model become inconsequential.
Hence, $\mathcal{L}_{\mathrm{MoM}}$ functions as a pseudo-binary classification, distinguishing between the $\mathcal{O}$-class and the entity classes to detect the misclassifications of $\mathcal{O}$-class as an entity class.
MoM learning enables independence from such factors as the number of class samples, task features, and the model, making it adaptable to similarly imbalanced tasks.

For the $n$-th sentence $\bm{x}^{(n)}$, the loss function $\mathcal{L_{\mathrm{sentence}}}$, when applying MoM learning, is written as follows:
\begin{equation}
\begin{split}
    \mathcal{L_{\mathrm{sentence}}}(\bm{y}^{(n)}, \bm{p}^{(n)}) = & \lambda \cdot \mathcal{L}(\bm{y}^{(n)}, \bm{p}^{(n)}) \\
    &+ (1 - \lambda) \cdot \mathcal{L}_{\mathrm{MoM}}(\bm{y}^{(n)}, \bm{p}^{(n)}),
\end{split}
\end{equation}
where $\lambda$ is a hyperparameter that controls the trade-off between $\mathcal{L}$ and $\mathcal{L}_{\mathrm{MoM}}$.
MoM learning simplifies weight adjustments compared to WCE with a single hyperparameter $\lambda$.
Finally, the model loss $\mathcal{L_{\mathrm{model}}}$ is minimized with the training labels for the entire dataset as $\bm{Y}$ and the prediction probabilities as $\bm{P}$:
\begin{equation}
    \mathcal{L_{\mathrm{model}}}(\bm{Y}, \bm{P}) = \frac{1}{N}  \sum_{n=1}^{N}\ \mathcal{L_{\mathrm{sentence}}}(\bm{y}^{(n)}, \bm{p}^{(n)}).
\end{equation}

MoM learning can also be considered a type of multitask learning~\cite{caruana1997multitask} that improves performance by learning several similar tasks simultaneously and has achieved good results against the data imbalance in various fields~\cite{zhang2018learning,kitada2019conversion,spangher2021multitask}.

\input{tables/tex/dataset}

%% file: figures/tex/ponti.tex
\begin{figure}[t]
    \centering
    \includegraphics[width=\linewidth]{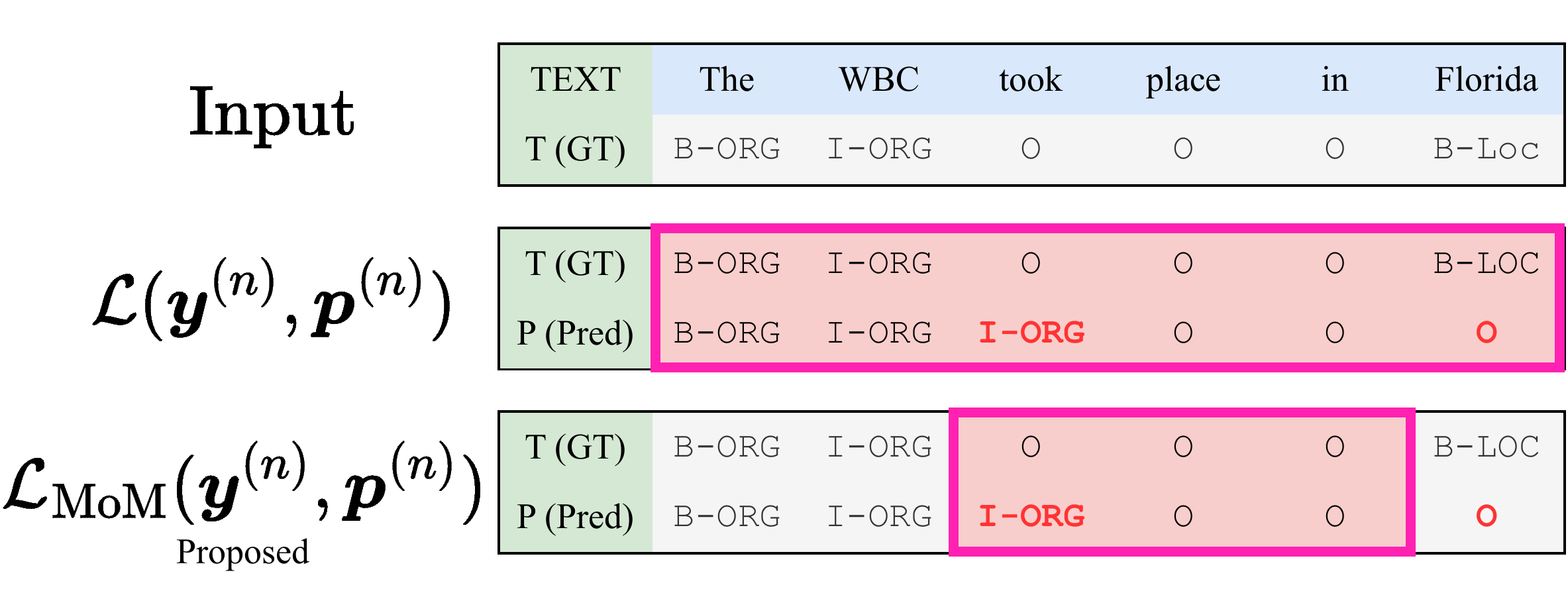}
    \caption{Concept of the MoM learning. The conventional loss function, $\mathcal{L}(\bm{y}^{(n)}, \bm{p}^{(n)})$ (e.g., cross-entropy loss), calculates the loss for all samples. In the MoM learning, $\mathcal{L}_{\mathrm{MoM}}(\bm{y}^{(n)}, \bm{p}^{(n)})$, the loss associated with the ``major'' $\mathcal{O}$-class is added to $\mathcal{L}(\bm{y}^{(n)}, \bm{p}^{(n)})$.}
    \label{fig:ponti}
\end{figure}

%% file: tables/tex/dataset.tex
\begin{table*}[ht]
\centering
\caption{Statistics for four dataset (English and Japanese). $\rho_{\mathcal{O}}$ indicates the proportion of $\mathcal{O}$-classes to the total. By dominant $\mathcal{O}$-class, $\rho_{\mathcal{O}}$ exceeds 80\% in each dataset.}
\begin{tabular}{@{}lcrrrrrrr@{}}
\toprule
                                        & \textbf{Lang.} & \textbf{\# train} & \textbf{\# val} & \textbf{\# test} & \textbf{\# class} & \textbf{\# $\mathcal{O}$-sample} & \textbf{\# entities sample} & $\rho_{\mathcal{O}}$ [\%] \\ \cmidrule(r){1-1} \cmidrule(lr){2-2} \cmidrule(lr){3-3} \cmidrule(lr){4-4} \cmidrule(lr){5-5} \cmidrule(lr){6-6} \cmidrule(lr){7-7} \cmidrule(lr){8-8} \cmidrule(l){9-9}
CoNLL2003~\cite{sang2003introduction}     & EN & 14,041 & 3,250 & 3,453 & 9 & 248,818 & 53,993 & 82.17           \\
OntoNotes5.0~\cite{pradhan2013towards} & EN & 75,187 & 9,603 & 9,479 & 37 & 1,441,685 & 190,310 & 88.34                                   \\
KWDLC~\cite{hangyo2012building}         & JP    & 12,836   & 1,602  & 1,613 & 17 & 236,290  & 16,694 & 93.40                                   \\
NER Wiki~\cite{ohmi2021ner}             & JP  & 4,274 & 535 & 534 & 17 & 80,944 & 17,552 & 82.18                                   \\ \bottomrule
\end{tabular}%
\label{tab:dataset}
\end{table*}

%% file: 04_experiments.tex
This section describes the evaluation experiments, including sequential labeling and MRC, followed by the datasets, loss functions, and implementation details.
Considering the data variability, the evaluation was based on the average of the results of ten random seeds in each condition.
As evaluation metrics, we compare precision (Prec.) and recall (Rec.), as well as the F1 score.
For the F1 score, we selected an appropriate metric for each of the following frameworks.
In all evaluations, we performed paired t-tests ($\alpha=0.05$) to identify differences between our method and other best methods where $\alpha$ is the significance level.

\subsection{NER Frameworks}
\paragraph{Sequential labeling framework}
The sequential labeling classifies at the token level and yields the data imbalance with a long-tail distribution.
This framework directly addresses NER as a multiclass classification.
Thus, we used the macro F1 score as an evaluation criterion.

\paragraph{MRC framework}
Compared to sequential labeling, the MRC, another practical framework for NER, has been widely used in binary classification tasks in recent years~\cite{li2020unified}.
This framework determines whether each word belongs to a particular class and finds its range.
Specifically, for a token $\bm{w}^{(n)}_i$ in a sentence $\bm{x}^{(n)}$, the ground truth can be written as $\bm{y}^{(n)}_i \in \{0, 1\}^{\mathcal{Y}}$, where $\mathcal{Y}$ is the set of entity and non-entity classes.
Different from the macro F1 score for sequential labeling, we used the macro F1 score, which matches the index of the predicted start and end points.

\input{tables/tex/parameter}

\subsection{Datasets}
We used four datasets, as shown in \Cref{tab:dataset}: English CoNLL2003~\cite{sang2003introduction}, English OntoNotes5.0~\cite{pradhan2013towards}, Kyoto University web document read corpus (KWDLC; in Japanese)~\cite{hangyo2012building}, and Stockmark NER wiki (NER wiki; in Japanese)~\cite{ohmi2021ner}.
In addition, we evaluated four datasets using sequential labeling.
For MRC, we used the CoNLL2003, which has been adopted in previous studies, by converting the data from sequential labeling annotations.
For the English datasets, we employed the standard training, validation, and test data provided, and for the Japanese datasets without the standard ones, we randomly split the data 8:1:1.
\#~$\mathcal{O}$-sample and \#~entities sample are presented the number of $\mathcal{O}$-class and entity classes, respectively.
The imbalance rate $\rho_{\mathcal{O}}$ is the ratio of the number of $\mathcal{O}$-class samples to the total number of samples.

\subsection{Loss Functions}
We compared the prediction performance of MoM learning with that of conventional learning methods (e.g., loss functions), which have long been considered state-of-the-art and used widely for data imbalance.

\paragraph{WCE}
The WCE is one of the most generally used weighted loss functions, and we consider the following variants: (1) the inverse class frequency (WCE-1) and (2) a hyperparameter related to the number of samples (WCE-2).
The WCE loss, $\ell_{\mathrm{WCE}}$ for any class $k$ of any one sentence $\bm{x}^{(n)}$ in the NER is defined as follows:
\begin{equation}
    \ell_{\mathrm{WCE}}(\bm{y}^{(n)}_{i}, \bm{p}^{(n)}_{i}) = - \sum_{k}^{\mathcal{Y}} \omega_{k} {y}^{(n)}_{ik} \log{p}^{(n)}_{ik}.
\end{equation}
As $\omega_{k}$ in our experiments, WCE-1 is set as the inverse class frequency, and WCE-2 is set as $\log_{10}(\frac{s-s_k}{s_k}+\beta)$, used in a DL paper~\cite{li2020dice}, where $s_k$ is the number of samples for class $k$, $s$ is the total number of train samples and $\beta$ is a hyperparameter.

\paragraph{FL} 
The FL~\cite{lin2017focal} is a more robust and versatile loss~\cite{iikura2021improving, liu2021relation} proposed later than WCE:
\begin{equation}
    \ell_{\mathrm{FL}}(\bm{y}^{(n)}_{i}, \bm{p}^{(n)}_{i}) = - \bm{y}^{(n)}_{i} (1-\bm{p}^{(n)}_{i})^{\gamma} \log\bm{p}^{(n)}_{i},
\end{equation}
where $\gamma$ is a hyperparameter to reduce the relative loss for well-classified samples.
Because FL was designed for binary classification, we extended it with a one-versus-the-rest method in sequential labeling.

\paragraph{DL}
The DL~\cite{li2020dice} was designed to reduce both false positives and false negatives and has long been considered a state-of-the-art method focused on MRC:
\begin{equation}
    \ell_{\mathrm{DSC}}(\bm{y}^{(n)}_{i}, \bm{p}^{(n)}_{i}) = \frac{2(1- {p}^{(n)}_{ik})^{\epsilon} {p}^{(n)}_{ik} \cdot {y}^{(n)}_{ik} + \delta}{(1- {p}^{(n)}_{ik})^{\epsilon} + {y}^{(n)}_{ik} + \delta},
\end{equation}
where $\epsilon$ and $\delta$ are hyperparameters of reducing the relative loss for well-classified samples and smoothing, respectively.

\input{tables/tex/bert}
\input{tables/tex/roberta}
\input{tables/tex/conll_ent}
\input{tables/tex/bert_mrc}
\input{tables/tex/length}

\subsection{Implementation Details}

\paragraph{Models}
We utilized pre-trained models, as shown in \Cref{tab:para}, where the input length of these models was determined by the maximum number of tokens in a sentence ($M=128$) with padding tokens (\texttt{[PAD]}) used for filling the remaining space to maintain a consistent length.
We fine-tuned each task in 10 epochs using the Adam optimizer~\cite{kingma2014adam}.

During sequential labeling, we used pre-trained BERT~\cite{devlin2019bert} and RoBERTa~\cite{liu2019roberta} as baseline models.
In the latter part of the $D$-dimensional special classification (\texttt{[CLS]}) token of these models, we attached a new head of $D \times (M \times \mathcal{Y})$ two-layered fully connected neural networks, and the head was trained to minimize the cross-entropy loss for each token.
As such, these models output the predicted probabilities of each class corresponding to the words in the text.
We set $D$ to 768, a learning rate of $2 \times 10^{-5}$ and a batch size of 64, respectively.

In MRC, we used BERT-MRC~\cite{li2020unified} as the baseline model, where the latter part of the \texttt{[CLS]} token, we used the same architecture as sequential labeling with changed $\mathcal{Y}$ to $2$.
As such, this model outputs a range corresponding to an entity for a sentence and $\mathcal{Y}$ binary classification.
We set a learning rate of $3 \times 10^{-5}$ and a batch size of 32, respectively.

\paragraph{Hyperparameters}Tree-Structured Parzen Estimator (TPE)~\cite{bergstra2011algorithms}, implemented in the Bayesian optimization library Optuna~\cite{akiba2019optuna}, to maximize the F1 score of the validation data. 
For the sequential labeling experiments, the hyperparameters of WCE-2 ($\beta$) and FL ($\gamma$) were explored in the pre-determined range of 1.0-10.0 and 0.0-10.0, respectively, considering their papers~\cite{lin2017focal, li2020dice}.
For the MRC experiments, we set the hyperparameters of FL $\gamma=3.0$, and those of DL $\epsilon=1.0$ and $\delta=0.01$, which is based on the DL hyperparameters carefully tuned in ~\cite{li2020dice}.
The hyperparameters of MoM ($\lambda$) were explored in the pre-determined range of 0.0-1.0 in both frameworks.
We show the results of tuning the MoM hyperparameter $\lambda$ in Table~\ref{tab:para}.

%% file: tables/tex/parameter.tex
\begin{table}[t]
    \caption{Summary of the pre-trained models for each hyper-parameter, where $\lambda$ is MoM learning for each dataset. We used Tree-Structured Parzen Estimator (TPE)~\cite{bergstra2011algorithms} implemented in a Bayesian optimization library Optuna~\cite{akiba2019optuna} according to the F1 score of the validation data. The part of the experiment not performed in this study is left blank.}
    \label{tab:para}
    
    \begin{minipage}{\linewidth}
        \centering
        \subcaption{English}
        \label{tab:para-english}
        
        \begin{tabular}{@{}llrr@{}}
            \toprule
            Model    & Pre-trained model & \multicolumn{1}{l}{CoNLL2003} & \multicolumn{1}{l}{OntoNotes5.0} \\ \cmidrule(r){1-1} \cmidrule(lr){2-2} \cmidrule(lr){3-3} \cmidrule(l){4-4}
            BERT     & BERT-based-cased\tablefootnote{\label{footnote:bert-base-cased}\url{https://huggingface.co/google-bert/bert-base-cased}}  & 0.175                         & 0.125                            \\
            RoBERTa  & xlm-roberta-base\tablefootnote{\url{https://huggingface.co/FacebookAI/xlm-roberta-base}}  & 0.209                         & 0.041                            \\
            BERT-MRC & BERT-base-cased\tablefootnote{\label{footnote:bert-base-cased}\url{https://huggingface.co/google-bert/bert-base-cased}}   & 0.446                         & ---                              \\ \bottomrule
        \end{tabular}
    \end{minipage}
    
    \vspace{2mm}
    
    \begin{minipage}{\linewidth}
        \centering
        \subcaption{Japanese}
        \label{tab:para-japanese}
        
        \begin{tabular}{@{}llrr@{}}
            \toprule
            Model   & Pre-trained model         & \multicolumn{1}{l}{KWDLC} & \multicolumn{1}{l}{NER Wiki} \\ \cmidrule(r){1-1} \cmidrule(lr){2-2} \cmidrule(lr){3-3} \cmidrule(l){4-4}
            BERT    & Tohoku Univ. BERT-base\tablefootnote{\url{https://huggingface.co/tohoku-nlp/bert-base-japanese}}    & 0.357                     & 0.212                        \\
            RoBERTa & Waseda Univ. RoBERTa-base\tablefootnote{\url{https://huggingface.co/nlp-waseda/roberta-base-japanese}} & 0.291                     & 0.248                        \\ \bottomrule
        \end{tabular}
    \end{minipage}
\end{table}

%% file: tables/tex/bert.tex
\begin{table*}[t]
    \centering
    \caption{The performance with BERT in the sequential labeling task (in macro F1).
    In all items, the proposed MoM had the best score, with a significant difference in FL, which was the next best ($\alpha=0.05$).}
    \label{tab:bert}
    \begin{tabular}{@{}lrrrrrrrrrrrr@{}}
        \toprule
        & \multicolumn{3}{l}{\textbf{CoNLL2003}} & \multicolumn{3}{l}{\textbf{OntoNotes5.0}} & \multicolumn{3}{l}{\textbf{KWDLC}} & \multicolumn{3}{l}{\textbf{Stockmark NER Wiki}} \\ \cmidrule(lr){2-4} \cmidrule(lr){5-7} \cmidrule(lr){8-10} \cmidrule(l){11-13}
        & \multicolumn{1}{l}{\textbf{Prec.}} & \multicolumn{1}{l}{\textbf{Rec.}} & \multicolumn{1}{l}{\textbf{F1}} & \multicolumn{1}{l}{\textbf{Prec.}} & \multicolumn{1}{l}{\textbf{Rec.}} & \multicolumn{1}{l}{\textbf{F1}} & \multicolumn{1}{l}{\textbf{Prec.}} & \multicolumn{1}{l}{\textbf{Rec.}} & \multicolumn{1}{l}{\textbf{F1}} & \multicolumn{1}{l}{\textbf{Prec.}} & \multicolumn{1}{l}{\textbf{Rec.}} & \multicolumn{1}{l}{\textbf{F1}} \\ \cmidrule(r){1-1} \cmidrule(lr){2-2} \cmidrule(lr){3-3} \cmidrule(lr){4-4} \cmidrule(lr){5-5} \cmidrule(lr){6-6} \cmidrule(lr){7-7} \cmidrule(lr){8-8} \cmidrule(lr){9-9} \cmidrule(lr){10-10} \cmidrule(lr){11-11} \cmidrule(lr){12-12} \cmidrule(l){13-13}
        BERT (baseline) & 90.16 & 91.86  & 91.00 & 87.41 & 89.07 & 88.23 & 70.92 & 73.96 & 72.41 & 77.32 & 81.04 & 79.13 \\ \cmidrule(r){1-1} \cmidrule(lr){2-2} \cmidrule(lr){3-3} \cmidrule(lr){4-4} \cmidrule(lr){5-5} \cmidrule(lr){6-6} \cmidrule(lr){7-7} \cmidrule(lr){8-8} \cmidrule(lr){9-9} \cmidrule(lr){10-10} \cmidrule(lr){11-11} \cmidrule(lr){12-12} \cmidrule(l){13-13}
        w/ WCE-1 & 89.73 & 92.15 & 90.93 & 85.66 & 90.28 & 87.91 & 62.79 & 78.32 & 69.70 & 73.65 & 80.28 & 76.82 \\
        \multicolumn{1}{r}{} &  &  & (-0.07) & & & (-0.32) & & & (-2.71) & & & (-2.31) \\
        w/ WCE-2 & 89.94 & 92.22 & 91.07 & 86.81 & 89.67 & 88.22 & 68.86 & 77.23 & 72.80 & 75.72 & 81.19 & 78.36 \\
        \multicolumn{1}{r}{} &  &  & (+0.07) & & & (-0.01) & & & (+0.39) & & & (-0.77) \\
        w/ FL & 90.33 & 92.03 & 91.17 & 87.62 & 89.15 & 88.39 & 71.88 & 74.27 & 73.05 & 77.79 & 81.53 & 79.61 \\
        \multicolumn{1}{r}{} & & & (+0.17) & & & (+0.16) & & & (+0.64) & & & (+0.48) \\ \cmidrule(r){1-1} \cmidrule(lr){2-2} \cmidrule(lr){3-3} \cmidrule(lr){4-4} \cmidrule(lr){5-5} \cmidrule(lr){6-6} \cmidrule(lr){7-7} \cmidrule(lr){8-8} \cmidrule(lr){9-9} \cmidrule(lr){10-10} \cmidrule(lr){11-11} \cmidrule(lr){12-12} \cmidrule(l){13-13}
        \textbf{w/ MoM} & 90.41 & 92.27 & \textbf{91.33} & 87.39 & 89.84 & \textbf{88.60} & 72.54 & 74.13 & \textbf{73.32} & 78.13 & 81.61 & \textbf{79.83} \\
        \multicolumn{1}{}{} (\textbf{proposed})& & & (+0.33) & & & (+0.37) & & & (+0.91) & & & (+0.70) \\ \bottomrule
    \end{tabular}
\end{table*}

%% file: tables/tex/roberta.tex
\begin{table*}[t]
\centering
\caption{The performance with RoBERTa in the sequential labeling task (in macro F1) .
In all items, the proposed MoM had the best score, with a significant difference in FL, which was the next best ($\alpha=0.05$) except for OntoNote $\dagger$}
\label{tab:roberta}
\begin{tabular}{@{}lrrrrrrrrrrrr@{}}
\toprule
                     & \multicolumn{3}{l}{\textbf{CoNLL2003}}                                                   & \multicolumn{3}{l}{\textbf{OntoNotes5.0}}                                              & \multicolumn{3}{l}{\textbf{KWDLC}}                                                     & \multicolumn{3}{l}{\textbf{Stockmark NER Wiki}}                                        \\ \cmidrule(lr){2-4} \cmidrule(lr){5-7} \cmidrule(lr){8-10} \cmidrule(l){11-13}
                     & \multicolumn{1}{l}{\textbf{Prec.}} & \multicolumn{1}{l}{\textbf{Rec.}} & \multicolumn{1}{l}{\textbf{F1}} & \multicolumn{1}{l}{\textbf{Prec.}} & \multicolumn{1}{l}{\textbf{Rec.}} & \multicolumn{1}{l}{\textbf{F1}} & \multicolumn{1}{l}{\textbf{Prec.}} & \multicolumn{1}{l}{\textbf{Rec.}} & \multicolumn{1}{l}{\textbf{F1}} & \multicolumn{1}{l}{\textbf{Prec.}} & \multicolumn{1}{l}{\textbf{Rec.}} & \multicolumn{1}{l}{\textbf{F1}} \\ \cmidrule(r){1-1} \cmidrule(lr){2-2} \cmidrule(lr){3-3} \cmidrule(lr){4-4} \cmidrule(lr){5-5} \cmidrule(lr){6-6} \cmidrule(lr){7-7} \cmidrule(lr){8-8} \cmidrule(lr){9-9} \cmidrule(lr){10-10} \cmidrule(lr){11-11} \cmidrule(lr){12-12} \cmidrule(l){13-13}
RoBERTa (baseline) & 89.93                     & 91.56                    & 90.74                  & 88.24                     & 90.00                    & 89.11                  & 77.11                     & 81.66                    & 79.31                  & 81.07                     & 84.45                    & 82.72                  \\ \cmidrule(r){1-1} \cmidrule(lr){2-2} \cmidrule(lr){3-3} \cmidrule(lr){4-4} \cmidrule(lr){5-5} \cmidrule(lr){6-6} \cmidrule(lr){7-7} \cmidrule(lr){8-8} \cmidrule(lr){9-9} \cmidrule(lr){10-10} \cmidrule(lr){11-11} \cmidrule(lr){12-12} \cmidrule(l){13-13}
w/ WCE-1            & 85.86                     & 91.31                    & 88.50                  & 59.55                     & 82.59                    & 69.20                  & 46.85                     & 77.38                    & 58.31                  & 39.51                         & 49.58                        & 43.98                  \\
\multicolumn{1}{r}{} &                           &                          & (-2.24)                &                           &                          & (-19.91)                &                           &                          & (-21.00)                &                           &                          & (-38.74)                \\
w/ WCE-2            & 89.88                     & 91.94                   & 90.90                & 86.61                    & 90.49                   & 88.51                 & 68.86                    & 77.23                   & 72.80                 & 75.72                        & 81.19                      & 78.36                 \\
\multicolumn{1}{r}{} &                           &                          & (+0.16)                &                           &                          & (-0.60)                &                           &                          & (-6.51)                &                           &                          & (-4.36)                \\
w/ FL            & 90.97                    & 91.11                   & 91.04                  & 88.35                   & 90.09                   & 89.22                  & 80.12                    & 82.44                   & 81.28                  & 81.41                        & 85.07                       & 83.24                  \\
\multicolumn{1}{r}{} &                           &                          & (+0.30)                &                           &                          & (+0.11)                &                           &                          & (+1.93)               &                           &                          & (+0.52)                \\ \cmidrule(r){1-1} \cmidrule(lr){2-2} \cmidrule(lr){3-3} \cmidrule(lr){4-4} \cmidrule(lr){5-5} \cmidrule(lr){6-6} \cmidrule(lr){7-7} \cmidrule(lr){8-8} \cmidrule(lr){9-9} \cmidrule(lr){10-10} \cmidrule(lr){11-11} \cmidrule(lr){12-12} \cmidrule(l){13-13}
\textbf{w/ MoM}           & 91.11                    & 91.27                  & \textbf{91.19}         & 88.34                   & 90.16                   & \textbf{89.25}$\dagger$         & 81.10                   & 82.60                   & \textbf{81.85}         & 81.85                   & 85.23                     & \textbf{83.54}         \\
\multicolumn{1}{}{} (\textbf{proposed})&                           &                          & (+0.45)                &                           &                          & (+0.14)                &                           &                          & (+2.54)                &                           &                          & (+0.82)                \\ \bottomrule
\end{tabular}
\end{table*}

%% file: tables/tex/conll_ent.tex
\begin{table}[t]
\centering
\caption{Comparison of performance in each entity in sequential labeling of the CoNLL2003 dataset.
The prefixes \texttt{B} and \texttt{I} are marged for display purposes.
}
\label{tab:conll_ent}
\begin{tabular}{@{}lrrrrrrr@{}}
\toprule
        \multicolumn{1}{l}{}      & \multicolumn{3}{l}{\textbf{w/ MoM}}                         & \multicolumn{3}{l}{\textbf{BERT}}       & \multicolumn{1}{l}{}     \\ \cmidrule(lr){2-4} \cmidrule(lr){5-7} \cmidrule(l){8-8}
       & \multicolumn{1}{l}{\textbf{Prec.}} & \multicolumn{1}{l}{\textbf{Rec.}} & \multicolumn{1}{l}{\textbf{F1}} & \multicolumn{1}{l}{\textbf{Prec.}} & \multicolumn{1}{l}{\textbf{Rec.}} & \multicolumn{1}{l}{\textbf{F1}} & \multicolumn{1}{l}{\textbf{Diff}} \\ \cmidrule(r){1-1} \cmidrule(lr){2-2} \cmidrule(lr){3-3} \cmidrule(lr){4-4} \cmidrule(lr){5-5} \cmidrule(lr){6-6} \cmidrule(lr){7-7} \cmidrule(l){8-8}
\texttt{MISC}  & 79.18                    & 84.58                   & \textbf{81.78}        & 79.21                    & 84.04                   & 81.54                             & +0.24                   \\
\texttt{LOC} & 92.91                    & 93.50                   & \textbf{93.20}        & 92.60                    & 93.49                   & 93.04                     & +0.16                   \\
\texttt{ORG}  & 89.87                    & 93.17                   & \textbf{91.54}        & 89.90                    & 92.70                   & 91.27           & +0.27                   \\
\texttt{PER}  & 97.66                    & 97.88                   & \textbf{97.77}                 & 97.76                    & 97.55                   & 97.65              & +0.12                  \\
$\mathcal{O}$ & 99.72                    & 99.28                   & \textbf{99.50}        & 99.69                    & 99.31                   & \textbf{99.50}                & 0.00     \\ \bottomrule            
\end{tabular}%
\end{table}

%% file: tables/tex/bert_mrc.tex
\begin{table}[t]
\centering
\caption{Summary of the performance of the MRC task on the CoNLL2003 dataset.
MoM learning was also significant in MRC compared to FL and DL, which are state-of-the-art methods.
}
\label{tab:bert_mrc}
\begin{tabular}{@{}lrrr@{}}
\toprule
                        & \multicolumn{3}{l}{\textbf{CoNLL2003}}           \\ \cmidrule(l){2-4}
                        & \multicolumn{1}{l}{\textbf{Prec.}}        & \multicolumn{1}{l}{\textbf{Rec.}}       & \multicolumn{1}{l}{\textbf{F1}}                                 \\ \cmidrule(r){1-1} \cmidrule(lr){2-2} \cmidrule(lr){3-3} \cmidrule(l){4-4}
BERT-MRC (baseline)           & \multicolumn{1}{r}{92.43} & \multicolumn{1}{r}{92.22} & \multicolumn{1}{r}{92.32}          \\ \cmidrule(r){1-1} \cmidrule(lr){2-2} \cmidrule(lr){3-3} \cmidrule(l){4-4}
BERT-MRC w/ FL           & \multicolumn{1}{r}{92.95} & \multicolumn{1}{r}{92.10} & \multicolumn{1}{r}{92.52}          \\
                        &                           &                           & \multicolumn{1}{r}{(+0.20)}                            \\ 
BERT-MRC w/ DL           & \multicolumn{1}{r}{92.69} & \multicolumn{1}{r}{92.43} & \multicolumn{1}{r}{92.56}          \\
                        &                           &                           & \multicolumn{1}{r}{(+0.24)}                            \\ \cmidrule(r){1-1} \cmidrule(lr){2-2} \cmidrule(lr){3-3} \cmidrule(l){4-4}
\textbf{BERT-MRC w/ MoM} & \multicolumn{1}{r}{92.99} & \multicolumn{1}{r}{92.51} & \multicolumn{1}{r}{\textbf{92.75}} \\
(\textbf{proposed}) &                           &                           & \multicolumn{1}{r}{(+0.43)}                            \\ \bottomrule
\end{tabular}
\end{table}

%% file: tables/tex/length.tex
\begin{table}[t]
\centering
\caption{
The performance comparison by word count on CoNLL2003 test data in sequential labeling.
\# words indicate the number of words in each data, and \# data indicates the number of data with that word count.
To evaluate the sentence-level, we show the average accuracy set for every 5 words for the bin, calculated based on matching the labels of all words.
}
\label{tab:length}
\begin{tabular}{@{}rrrrrr@{}}
\toprule
& & & \multicolumn{3}{c}{\textbf{sentence-level Acc.}} \\ \cmidrule(lr){4-6}
\textbf{\# words} & \textbf{\# data} & $\rho_{\mathcal{O}}$ [\%] & \textbf{BERT} & \textbf{w/ FL} & \textbf{w/ MoM}  \\ \cmidrule(r){1-1} \cmidrule(lr){2-2} \cmidrule(lr){3-3} \cmidrule(lr){4-4} \cmidrule(l){5-5} \cmidrule(l){6-6} 
1-5 & 874 & 64.32 & 89.36 & 89.17 & \textbf{89.79}  \\        
6-10 & 1,190 & 77.59 & 87.22  & \textbf{88.26} & 87.81   \\
11-15  & 310 & 80.05 & 81.43  & 82.15 & \textbf{82.58}  \\
16-20  & 244 & 89.12 & 88.01  & 88.49 & \textbf{89.69} \\        
21-25  & 254 & 87.12 & 83.02  & 84.44 & \textbf{84.55} \\
26-30  & 204 & 87.02 & \textbf{80.05}  & 79.74 & 79.62 \\
31-35  & 175 & 86.65 & 72.99 & 75.78 & \textbf{76.50} \\
36-  & 202 & 75.63 & 85.98 & 86.12 & \textbf{86.56} \\ \bottomrule
\end{tabular}
\end{table}

%% file: 05_results.tex
\input{figures/tex/results}

Tables~\ref{tab:bert} and \ref{tab:roberta} present a comparison of the performances of each method using BERT and RoBERTa in sequential labeling.
We confirmed that MoM learning consistently outperforms other tested methods across all four datasets, regardless of these models.
In all results (4 datasets $\times$ 2 models), the performance using MoM learning was significant at $\alpha=0.05$ against the other best method (FL).
In the OntoNotes5.0 results with RoBERTa (marked as $\dagger$), MoM learning was slightly less significant, but it was significant compared to the baseline CE.
The results using WCE-1 and WCE-2 demonstrated poor performance compared to the baseline, and the reasons for this are discussed in \Cref{sec:discussion}.

\Cref{tab:conll_ent} compares the baseline with/without MoM learning for each entity in the CoNLL2003.
The prefixes of the entity classes \texttt{B} and \texttt{I} are merged to show the average performance of the respective classes, resulting in 9 classes (e.g., \texttt{B-PER}, \texttt{I-PER}, \texttt{B-LOC} and $\mathcal{O}$) becoming 5 classes (e.g., \texttt{PER}, \texttt{LOC}, and $\mathcal{O}$).
We confirm that MoM learning improves the performance of entity classes without compromising the performance of the $\mathcal{O}$-class.

\Cref{tab:bert_mrc} presents the performance with the CoNLL2003 dataset using the MRC. 
The results confirm MoM learning also demonstrated the best performance and was significant at $\alpha=0.05$, against the other best method DL.

\Cref{tab:length} shows CoNLL2003 test performance in sequential labeling by word count. For sentence-level evaluation, the score is based on accuracy matching all word labels in the sentence, rather than conventional macro-F1. To reduce data variability, the bin is set to 5 words, and average scores are calculated. Thus, MoM learning improves performance regardless of word count compared to baseline CE and FL.


\Cref{fig:enter-label} presents the results of evaluating the performance with different numbers of training data on the CoNLL2003 regarding macro-F1, sentence-level accuracy, and word-level accuracy, respectively.
Since the sampling methods can not be adapted while preserving the distribution of labels, we use random undersampling to reduce the dataset to 3/4, 2/3, 1/2, 1/3, 1/10, 6/100, and 3/100.
In those evaluations, MoM learning showed the best results even under low resources, regardless of the number of training data.

%% file: figures/tex/results.tex
\begin{figure}[t]
    \begin{minipage}{\linewidth}
        \centering
        \includegraphics[width=\linewidth]{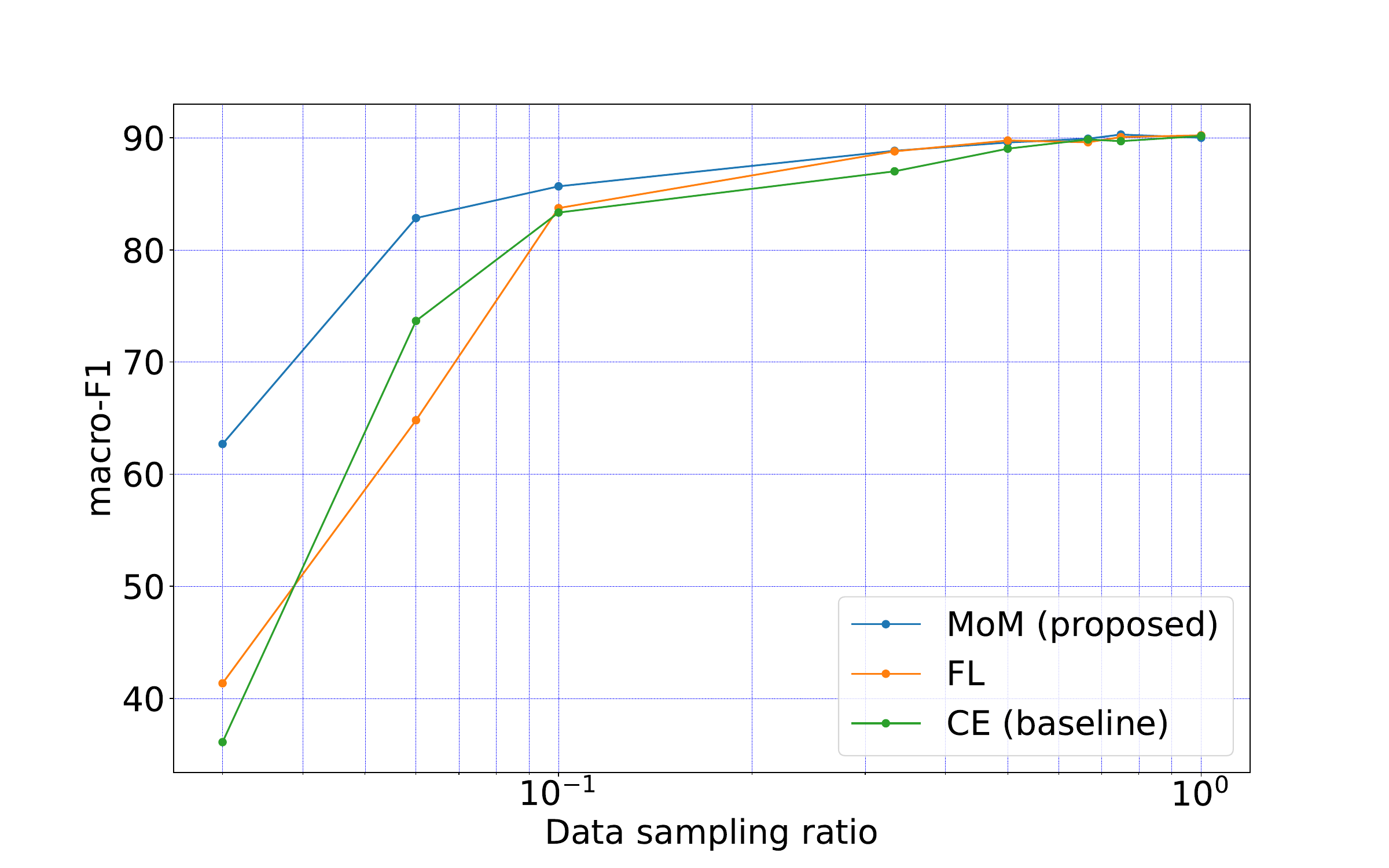}
        \subcaption{macro-F1}
        \label{fig:results_f1}
    \end{minipage}
    \begin{minipage}{\linewidth}
        \centering
        \includegraphics[width=\linewidth]{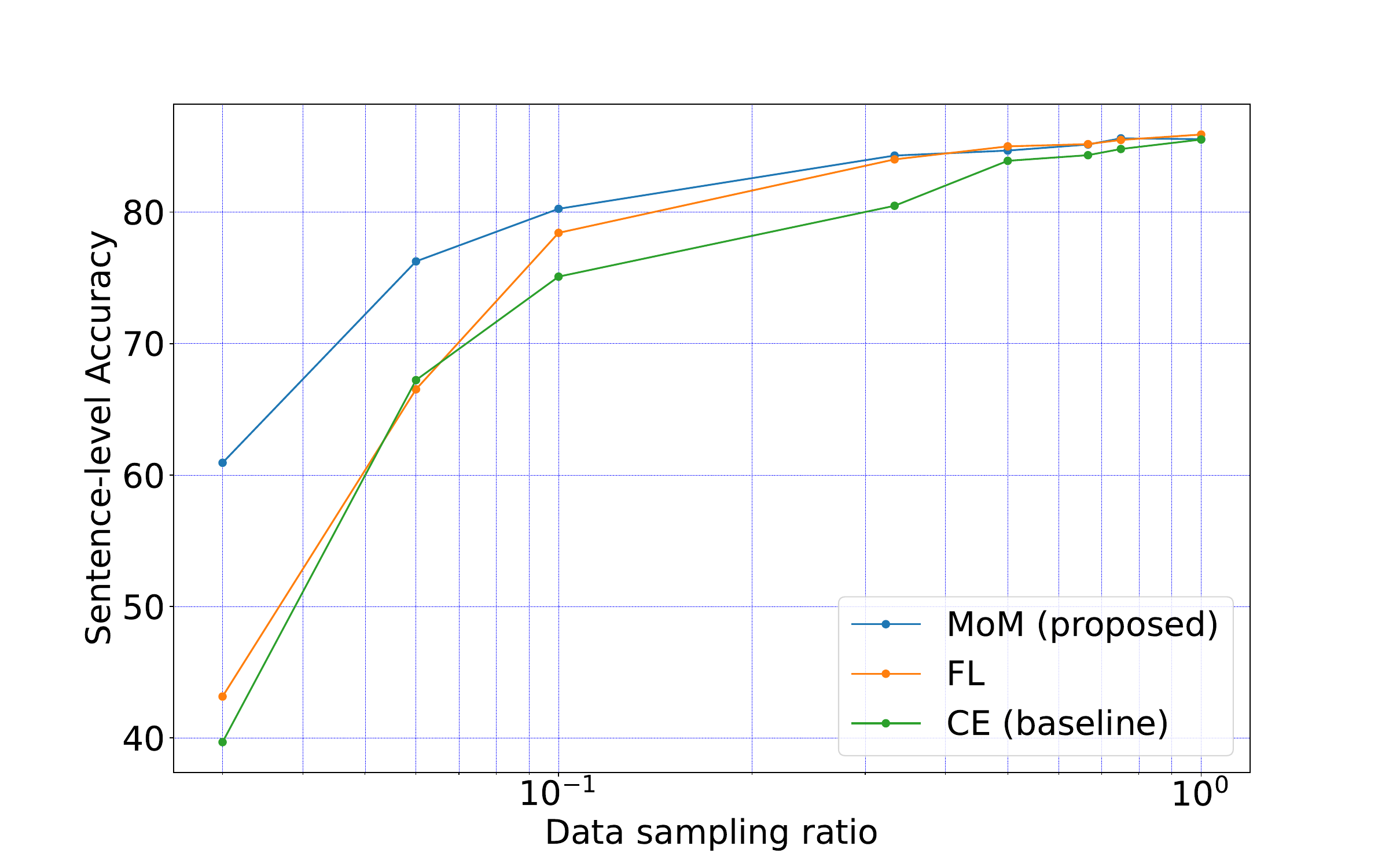}
        \subcaption{sentence-level}
        \label{fig:results_sentence-level}
    \end{minipage}
    \begin{minipage}{\linewidth}
        \centering
        \includegraphics[width=\linewidth]{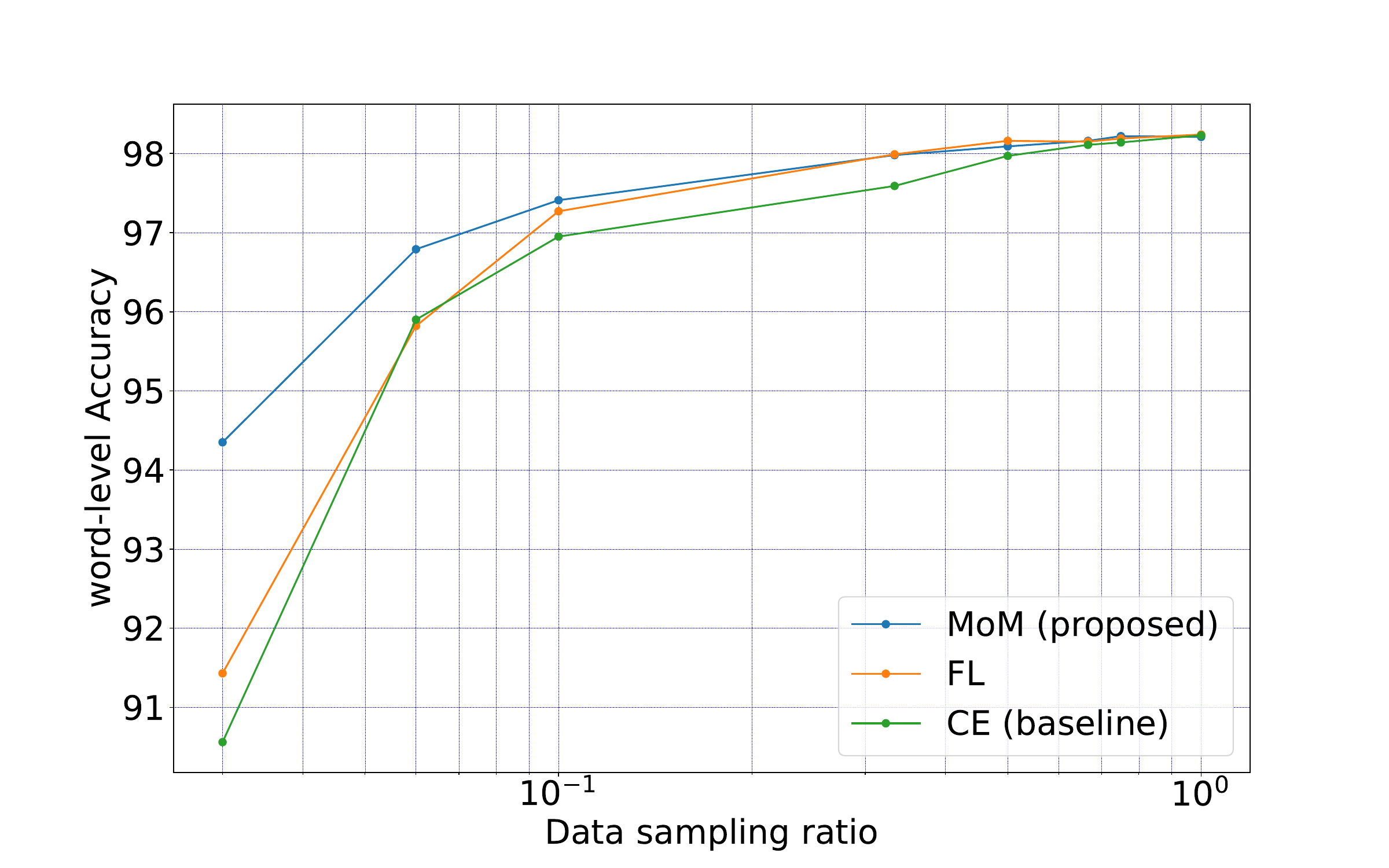}        
        \subcaption{word-level}
        \label{fig:results_word-level}
    \end{minipage}
    \caption{The result of evaluating the performance with different numbers of training data on the CoNLL2003. Since the sampling methods can not be adapted while preserving the distribution of labels, we use random undersampling to reduce the dataset to 3/4, 2/3, 1/2, 1/3, 1/10, 6/100, and 3/100. MoM learning showed the best results even under low resources, regardless of the number of training data.}
    \label{fig:enter-label}
\end{figure}

%% file: 06_discussion.tex
The most important factor in NER is the score of the entity classes, rather than the overall score, including $\mathcal{O}$-class, as the prediction performance of NER generally concerns the score including the $\mathcal{O}$-class.
In practical situations in which entities are extracted and utilized, the performances of the entity classes hold greater significance.
Although MoM learning appears a marginal improvement, we confirm MoM learning improves the performance of minor entity classes without sacrificing the performance of the major $\mathcal{O}$-class, regardless of the language and models.

For WCE, we attempted two methods (i.e., WCE-1 and WCE-2); however, a poor performance compared to the baseline CE was observed, highlighting the challenges posed by multiclass NER with its inherent long-tail distribution.
As evidenced by various ML tasks, conventional weighting methods struggle with the delicate design of loss functions dependent on specific datasets and tasks~\cite{valverde2017improving, jadon2020survey}.
Thus, the experiments in Tables~\ref{tab:bert} and \ref{tab:roberta} highlight the difficulty of applying WCE weighting to the sequential labeling of NER.

MoM learning was effective at sequential labeling and MRC, especially the latter, where we observe the role of MoM learning in monitoring the number of entities.
For example, in the sentence ``Estadio Santiago Bernabéu opened in 1974.'', ``Estadio'', ``Santiago'' and ``Bernabéu'' are assigned the classes \texttt{B-LOC}, \texttt{I-LOC}, \texttt{I-LOC}, and other words are assigned the $\mathcal{O}$-class, respectively.
when considering a basic sentence, it is highly likely that the word (``opened'') following the last word (``Bernabéu'') of the entity belongs to \texttt{I-LOC} or $\mathcal{O}$-class because sequences of different entity words are extremely rare, such as \texttt{LOCATION} after \texttt{PERSON}.
Because MoM learning focuses more on $\mathcal{O}$-class words, the model can learn whether the final word belongs to the \texttt{I-LOC} or $\mathcal{O}$-class.
In other words, the MoM can monitor the number of consecutive entity words, which is a factor in the improved performance of the words at the end of the entity.



Compared to FL, MoM learning improved performance on each score even when data was sampled, as shown in \Cref{fig:enter-label}.
In addition, MoM learning improved performance more as the data size decreased.
This contribution is because FL needs to set a hyperparameter $\gamma$ that depends on the difficulty of classification.
Since sampling methods produce variations in classification difficulty, setting the hyperparameter appropriately was difficult and could not have contributed to the performance.
Therefore, MoM learning consistently contributed to the performance improvement because it did not depend on the difficulty of classification, and $\rho_{\mathcal{O}}$ was almost the same regardless of the word count.
In addition, since MoM learning contributes to sentence-level accuracy regardless of data size, when we assume an application, this contribution is expected to reduce human load, such as rechecking by annotators.

MoM learning is successful in that it does not depend on the word count or data size, but there is still room for improvement.
Since MoM learning focuses on misclassification between $\mathcal{O}$ classes and entity classes, more is needed for misclassification between entity classes.
Further improving the prediction performance is possible by introducing a method specialized for entity classes.

%% file: 07_conclusion.tex

In this paper, we proposed a novel learning method, MoM learning, to tackle the NER task characterized by data imbalance with a long-tail distribution, consisting of a single class with many samples (the majority class) and multiple classes with few samples (the minority classes). MoM learning is a simple and effective method that reduces misclassifications of majority as minority classes by incorporating the loss of samples whose ground truth is the majority class into the loss of conventional ML models. Evaluation experiments using four datasets (English and Japanese) showed that MoM learning outperforms existing and even state-of-the-art methods in addressing data imbalances regardless of language, model, data size, or framework, whether sequential labeling or MRC.